%% file: main.tex
\documentclass{article}

\usepackage[preprint]{neurips_2022}

\usepackage[utf8]{inputenc} %
\usepackage[T1]{fontenc}    %
\usepackage{hyperref}       %
\usepackage{url}            %
\usepackage{booktabs}       %
\usepackage{amsfonts}       %
\usepackage{nicefrac}       %
\usepackage{microtype}      %
\usepackage{xcolor}         %
\usepackage{lipsum}
\usepackage{comment}

\usepackage{graphicx}
\input{math_commands.tex}

\title{Finding NEEMo: Geometric Fitting using Neural Estimation of the Energy Mover’s Distance}

\author{Ouail Kitouni, Niklas Nolte, Mike Williams\\
The NSF AI Institute for Artificial Intelligence and Fundamental Interactions\\
Massachusetts Institute of Technology\\
Cambridge, MA 02139, USA \\
\texttt{\{kitouni,nnolte,mwill\}@mit.edu} \\
}

\begin{document}

\maketitle

\begin{abstract}
A novel neural architecture was recently developed that enforces an exact upper bound on the Lipschitz constant of the model by constraining the norm of its weights 
in a minimal way, resulting in higher expressiveness compared to other techniques.
We present a new and interesting direction for this architecture:  estimation of the Wasserstein metric (Earth Mover’s Distance) in optimal transport by employing the Kantorovich-Rubinstein duality to enable its use in geometric fitting applications. %
Specifically, we focus on the field of high-energy particle physics, where it has been shown that a metric for the space of particle-collider events can be defined based on the Wasserstein metric, referred to as the Energy Mover's Distance (EMD). %
This metrization has the potential to revolutionize data-driven collider phenomenology. %
The work presented here represents a major step towards realizing this goal by providing a differentiable way of directly calculating the EMD. 
We show how the flexibility that our approach enables can be used to develop novel clustering algorithms. %
\end{abstract}

\section{Introduction}
The Wasserstein (Earth Mover's) Distance is a metric defined between two probability measures.
In the field of high-energy particle physics, a modified version of the Wasserstein distance, the Energy Mover's Distance (EMD), serves as a metric for the space of collider events by defining the {\em work} required to rearrange the radiation pattern of one event into another~\cite{Komiske2019emd}.
In particular, the EMD is intimately connected to the structure of {\em infrared- and collinear-safe} 
observables used in the ubiquitous task of clustering particles into {\em jets} \cite{Komiske:2020qhg}, 
and is foundational in the SHAPER tool for developing geometric collider observables~\cite{SHAPER}.

\begin{figure}[t]
    \centering
\includegraphics[width=\textwidth]{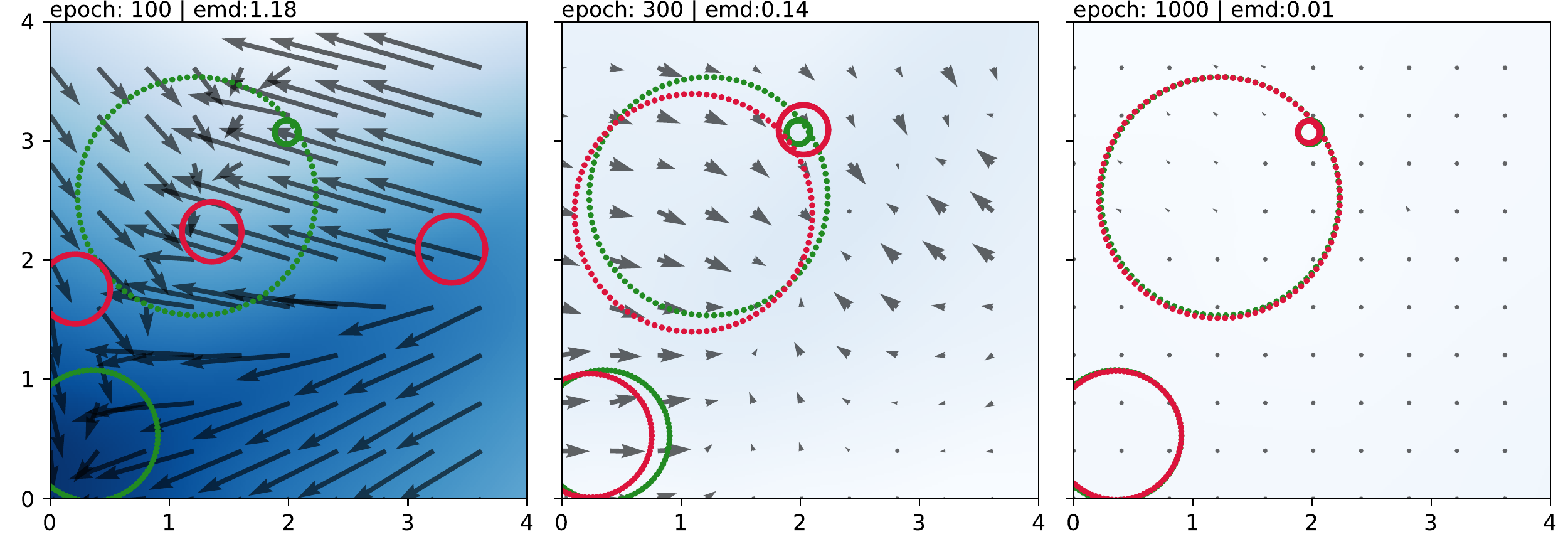}
\caption{Fitting three synthetic clusters (green) with three circles (red) using NEEMo (see Sec.~\ref{sec:NEEMo}). The heatmap is the Kantorovic potential, parameterized as a Lipschitz-bounded network, which induces forces on the circles (shown as arrows)
that drive them into perfect alignment with the target distribution
(only a few steps in the evolution of the fit are shown). 
\label{fig:circlefit}}
\end{figure}

Recently, a novel neural architecture was developed that enforces an exact upper bound on the Lipschitz constant of the model by constraining the norm of its weights 
in a minimal way, resulting in higher expressiveness than other methods~\cite{kitouni2021robust,sorting2019}.
Here, we employ this architecture---leveraging its improved expressiveness for 1-Lipschitz continuous networks---to replace the $\epsilon$-Sinkhorn estimation of the EMD in SHAPER~\cite{https://doi.org/10.48550/arxiv.1810.08278,SHAPER} by directly calculating the EMD using the  Kantorovic-Rubenstein (KR) dual formulation of the Wasserstein-1 metric. 
The KR duality casts the optimal transport problem as an optimization over the space of 1-Lipschitz functions, which we parameterize with dense neural networks using the architecture from~\cite{kitouni2021robust}.
With small modifications to the KR dual formulation, we are able to reliably and accurately obtain the EMD and Kantorovic potential in a differentiable way, without any $\epsilon$ approximations.
This makes it possible to run gradient-based optimization procedures over the exact EMD (see Fig.~\ref{fig:circlefit}). 
In addition, we expect these improvements could potentially have a major impact on jet studies at the future Electron-Ion Collider,  where traditional clustering methods are not optimal~\cite{Arratia:2020ssx}, and more broadly in  optimal transport problems.

\section{Lipschitz Networks and the Energy Mover's Distance}

\paragraph{Lipschitz Networks} Fully connected networks can be Lipschitz bounded by constraining the matrix
norm of all weights \cite{kitouni2021robust, p_norm_paper, spectral2018}.
Constraints with respect to a particular $\normlp$ norm will be denoted as Lip$^p$.
We start with a model $f(\vx)$ that is Lip$^p$ with Lipschitz constant $\lambda$ \emph{i.e.}, $\forall \, \vx, \bm{y} \in \mathbb{R}^n$: %
\begin{align} \label{eqn:sigma_lipschitz}
|f(\vx) - f(\bm{y})| &\leq \lambda\|\vx - \bm{y}\|_{p} \,.
\end{align}
Without loss of generality, we take $\lambda=1$ (rescaling the inputs would be equivalent to changing $\lambda$). We recursively define the layer $l$ of the fully connected network of depth $D$ with activation $\sigma$ as
\begin{align}
\label{eqn:gx}
\vz^l= \mW^l \sigma(\vz^{l-1}) + \vb^l ,
\end{align}
where $\vz^0 = \vx$ is the input and $f(\vx) = z^D$ is the output of the neural network. We have that $f(\vx)$ satisfies \eqref{eqn:sigma_lipschitz} if
\begin{align} \label{eqn:matrix_norm_sigma}
\|\mW^i\|_{\infty,\infty} \leq 1 \quad \mathrm{when} \quad 2\leq i\leq D \quad \mathrm{and} \quad ||\mW^1||_{p, \infty} \leq 1
\end{align}
and $\sigma$ has a Lipschitz constant less than or equal to 1. Here, $||\mW||_{p,q}$ denotes the operator norm with norm $L^p$ in the domain and $L^q$ in the co-domain. It is shown in \cite{sorting2019} that when using the \textbf{GroupSort} activation, $f(\vx)$ can approximate any Lip$^p$ function arbitrarily well, making weight-normed networks universal approximators. 
An implementation of the weight constraint along with a number of examples is provided in \href{https://github.com/niklasnolte/MonotOneNorm}{https://github.com/niklasnolte/MonotOneNorm}.

\paragraph{Energy Mover's Distance} The EMD is a metric between probability measures $\mathbb{P}$ and $\mathbb{Q}$. 
Using the standard Wasserstein-metric notation, the EMD is defined as 
\begin{align}
\mathrm{EMD}(\mathbb{P}, \mathbb{Q}) = \inf_{\gamma\in\Pi(\mathbb{P},\mathbb{Q})} \mathbb{E}_{(x,y)\sim\gamma} \big[||x-y||_2 \big] ,
\end{align}
where $\Pi(\mathbb{P},\mathbb{Q})$ is the set of all joint probability measures whose marginals are $\mathbb{P}$ and $\mathbb{Q}$. 
The Wasserstein optimization problem can be cast as an optimization over Lipschitz continuous functions using the Kantorovich-Rubinstein duality:
\begin{align}\label{eqn:kr_emd}
\mathrm{EMD}(\mathbb{P}, \mathbb{Q}) = \sup_{||f||_{L\leq1}} \mathbb{E}_{x\sim \mathbb{P}} \big[f(x)\big] - \mathbb{E}_{x\sim \mathbb{Q}} \big[f(x)\big], 
\end{align}
where $f$ is $\mathrm{Lip}^2$ continuous, \textit{i.e.,} $||\nabla f||_2 \leq 1$.
In high-energy particle collisions, the EMD is defined by using the energies of individual particles in place of probabilities, with their momentum directional coordinates representing the supports of the probability distribution.
For more details, including on how unequal total energies are handled,  
see~\cite{Komiske2019emd}. By performing optimizations over a constrained set of $\mathbb{P}$s, one can use the EMD to define observables over $\mathbb{Q}$. 

\begin{figure}[h!]
    \vspace*{-10pt}
    \centering
    \includegraphics[width=1\textwidth]{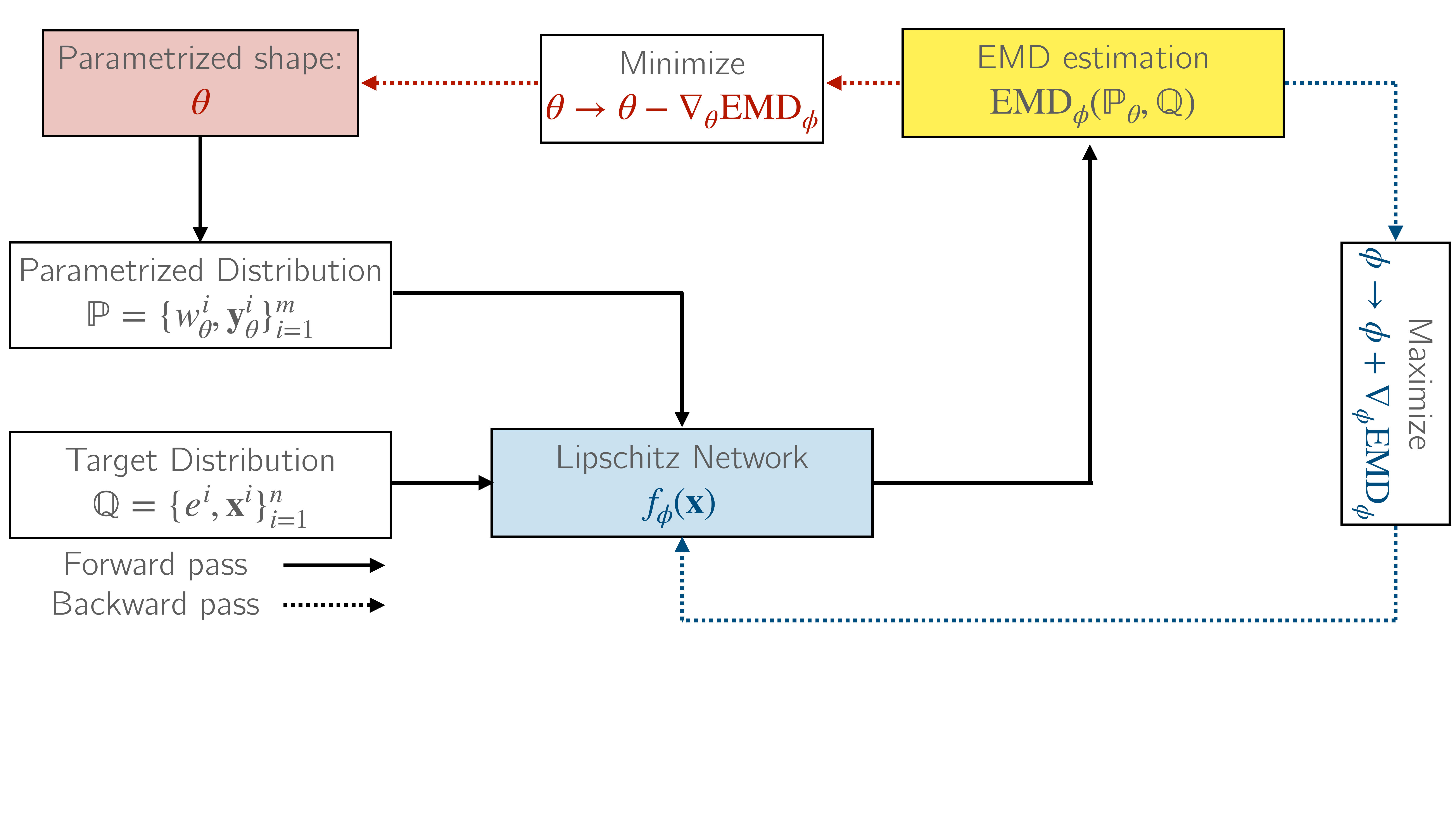}
    \vspace*{-60pt}
    \caption{Training procedure to fit a parameterized shape $\sP_\vtheta$ to a distribution $\sQ$. NEEMo replaces the $\epsilon$-Sinkhorn estimation in the standard SHAPER procedure with a Lipschitz network that evaluates the Kantorovic potential to obtain the EMD.}
    \label{fig:emd_diag}
\end{figure}

\section{NEEMo: Neural Estimation of the Energy Mover's Distance}
\label{sec:NEEMo}

\paragraph{Algorithm}
Consider a high-energy particle-collision event with $n$ particles. Let $E^i$ be the energy of particle $i$, $\vx^i$ be the direction of its momentum, and  $\sQ=\{(E^i,\vx^i) \}_{i=1}^n$ be the set of all particles in the event. 
Following the SHAPER prescription~\cite{SHAPER} for defining an observable $O(\sQ)$, we first define $\sP_\vtheta = \{w_\vtheta^i, \vy_\vtheta^i\}_{i=1}^m$ to be any collection of points parameterized by $\vtheta$, {\em e.g.}, these
points can be sampled from any geometric object with any density distribution. %
The EMD between the event $\sQ$ and the geometric object $\sP_\vtheta$ 
can be computed with \eqref{eqn:kr_emd} as
\begin{align}
    \label{eq:emdopt}
    \mathrm{EMD}(\sP_\vtheta, \sQ) = \max_{\vphi} \left[ \sum_{i=1}^{n}E^i f_\vphi(\vx^i) -\sum_{i=1}^{m}w_\vtheta^i f_\vphi(\vy_\vtheta^i) \right],
\end{align}
where $f_\vphi(x)$ is a 1-Lipschitz neural network with parameters $\vphi$. 
At $\vphi^*$ the expression above is maximized and $f_{\vphi^*}$ is the Kantorovic potential from which the EMD is obtained as the RHS of ~\eqref{eq:emdopt}.  
Since $f$ is differentiable, the optimum can be obtained using standard gradient descent techniques. This is the key improvement of NEEMo over SHAPER, which can only estimate the Kantorovic potential and the EMD up to a specified order $\epsilon$.
Note that in \eqref{eq:emdopt} the expectation is computed exactly but optimization can also be done stochastically by sampling from the discrete distributions with probabilities $\{E^i\}_i$ and $\{w_\theta^i\}_i$ and using the empirical mean to estimate the EMD. This can improve convergence in some cases.

Given that all of our operations are differentiable, gradients can flow back to $\sP_\vtheta$. Therefore, one can also optimize the parameters $\vtheta$ to obtain the best-fitting collection of points in that class. 
We obtain the following minimax optimization problem:  
\begin{align}
\label{eq:O}
    O(\sQ) = \min_\vtheta\max_\vphi 
    \left[ \sum_{i=1}^{n}E^i f_\vphi(\vx^i) -\sum_{i=1}^{m}w_\vtheta^i f_\vphi(\vy_\vtheta^i) \right],
\end{align}
where $O(\mathbb{Q})$ quantifies how well the event $\sQ$ is described by the class of geometric object $\sP$~\cite{Komiske:2020qhg,SHAPER}.

\paragraph{Limitations}
Unlike the conventional clustering algorithms used in high-energy particle physics, NEEMo relies on nonconvex gradient-based optimization of a neural network and a set of geometric parameters. 
This results in the clustering procedure itself being relatively slow and not easily implemented in real time. This problem can be alleviated with powerful custom optimizers and initialization techniques to guarantee fast convergence, though whether NEEMo could ever be run online during data taking is an open question. 
We note that for many potential applications, {\em e.g.}\ at the Electron-Ion Collider, this is not a problem since running online is not required. 

\section{Experiments}

\begin{figure}
    \centering
\includegraphics[width=\textwidth]{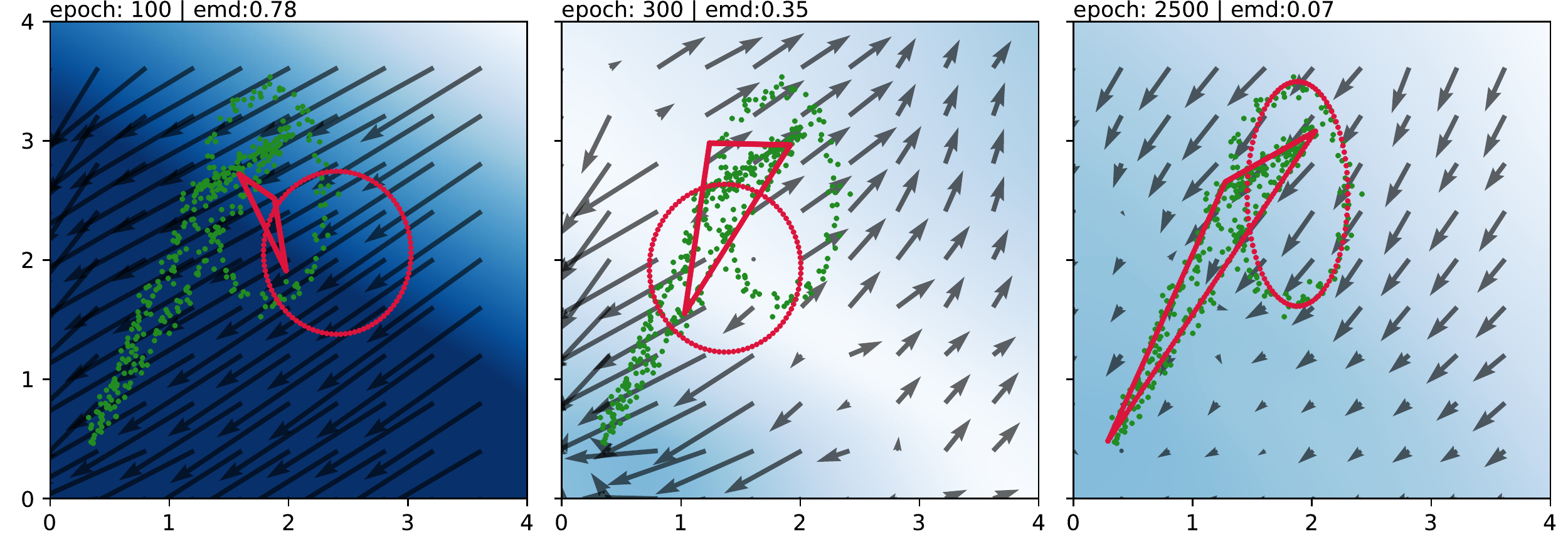}
\caption{Same as Fig.~\ref{fig:circlefit} but fitting to 
distributions parameterized by a triangle and an ellipse.
\label{fig:crazyfit}}
\end{figure}

\vspace{-0.5em}
\paragraph{Synthetic Data}
We start with a few toy examples.  
First, consider an event consisting of three sets of particles distributed uniformly along the perimeters of circles. 
Here, we know the exact parameterization of our target distribution and use NEEMo to fit three randomly initialized circles to the event. 
Figure~\ref{fig:circlefit} shows a few steps in the fit evolution. 
The Kantorovic potential given by the Lipschitz-constrained network induces forces on the parameters of $\sP$, which drive it to evolve from its random initialization to perfect alignment with the target distribution. 
In this example, $O(\sQ)$ in \eqref{eq:O} quantifies the {\em 3-circliness} of the event $\sQ$, an observable first defined in~\cite{SHAPER}.
To highlight the flexibility, %
we next consider an event with two sets of particles distributed along the perimeters of a triangle and ellipse, respectively.
Figure~\ref{fig:crazyfit} shows that $\sP$ again evolves following the gradients of the Kantorovic potential to perfect alignment with the target distribution.

\begin{figure}[h]
    \centering
    \includegraphics[width=0.9\textwidth]{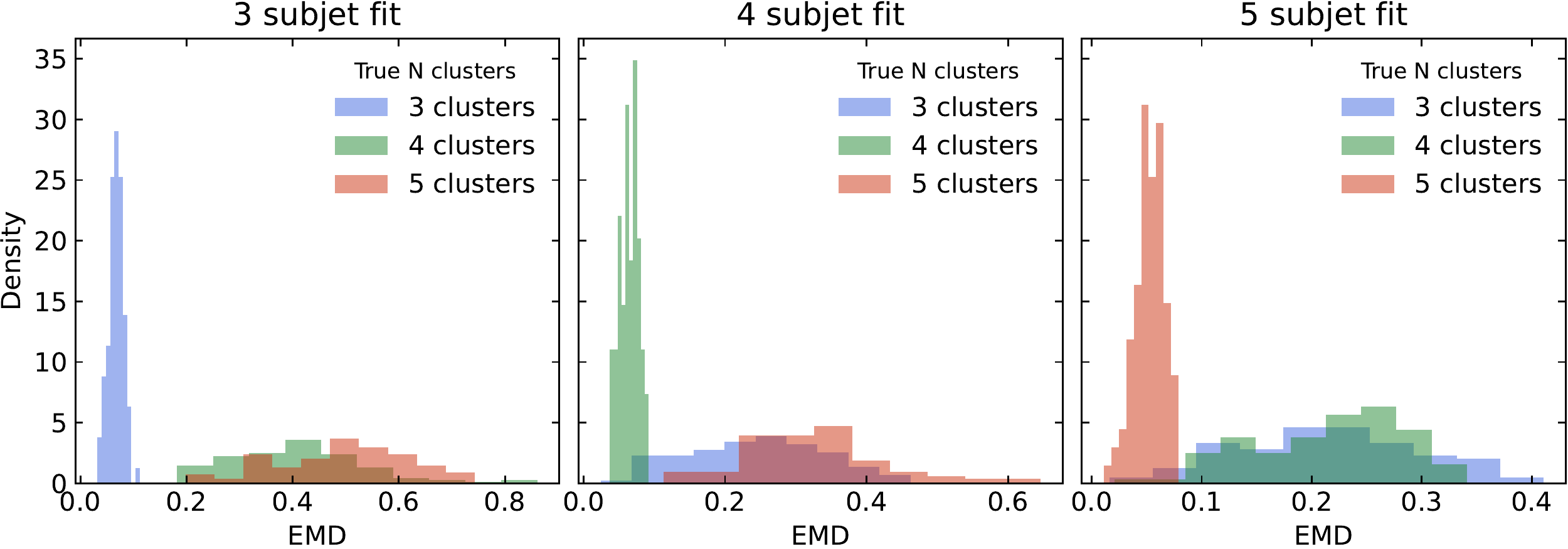}
    \caption{From left to right: Fit of N subjets (centers) to jets with 3, 4, or 5 number true subjets.}
    \label{fig:subjet}
\end{figure}

\vspace{-0.5em}
\paragraph{N-Subjets} 
We now perform a model jet-substructure study, clustering synthetic data into $N$-{\em subjets}. 
First, we generate jets with 3, 4, or 5 subjet centers distributed uniformly. 
From each center we generate 10 %
particles drawn from a Gaussian distribution. %
We then use our algorithm to fit 3, 4, or 5 centers to the simulated jets. 
Figure~\ref{fig:subjet} shows that our algorithm is able to estimate the correct number of subjets. 
The EMD of the N-subjet fit is clearly lowest for jets with N true clusters. %

\vspace{-0.5em}
\paragraph{Future Directions}
In the framework developed in these proceedings, any parameterized source distribution can be chosen to fit any target distribution using the EMD, without any $\epsilon$-approximations. This can be used, {\em e.g.}, for constructing precision jet observables that are sensitive to percent-level fluctuations for new physics searches at LHC experiments.
In addition, NEEMo provides a more precise way to quantify event modifications due to hadronization and detector effects.
Finally, the flexibility provided by NEEMo could potentially have a major impact on jet studies at the future Electron-Ion Collider,  where traditional clustering methods are not optimal. 
Rather than modifying the metric used in a sequential-recombination algorithm as in \cite{Arratia:2020ssx}, the jet geometry itself can be altered using NEEMo in an event-by-event unsupervised manner. 
We plan to report on all of these novel directions in a follow-up journal article that is currently in preparation.

\clearpage

\section{Broader Impacts}

Comparing probability distributions is a fundamental task in statistics. 
Most commonly used methods only compare densities in a point-wise manner, whereas the Earth Mover's Distance accounts for the geometry of the underlying space. 
This is easily visualized in our figures showing the Kantorovic potential. 
Due to space constraints we only showed a few toy example applications in collider physics, but we stress that the approach we present here---directly calculating the Earth Mover's Distance using the Kantorovic-Rubenstein dual formulation of the Wasserstein-1 metric---can be applied to any optimal transport problem. 
While the existence of the KR duality has long been known, it only recently became possible to simultaneously enforce the \textit{exact} 1-Lipschitz bound while achieving enough expressiveness to find the optimal  Kantorovic potential. 
Our approach now makes it possible to perform gradient-based optimizations over the exact Earth Mover's Distance. 
Given the sizable impact of similar approximate methods, we expect our exact approach could have applications across many fields and types of problems.

\section*{Acknowledgements}

We thank Rikab Gambhir and Jesse Thaler for helpful discussions about SHAPER, and for introducing us to this problem. 
This work was supported by NSF grant PHY-2019786 (The NSF AI Institute for Artificial Intelligence and Fundamental Interactions, http://iaifi.org/) and by DOE grant DE-FG02-94ER40818.

\bibliography{main}
\bibliographystyle{unsrtnat}

\end{document}

%% file: math_commands.tex
\usepackage{amsmath,amsfonts,bm}

\def\eqref#1{equation~\ref{#1}}

\def\1{\bm{1}}

\def\vtheta{{\bm{\theta}}}
\def\vphi{{\bm{\phi}}}

\def\vb{{\bm{b}}}

\def\vx{{\bm{x}}}
\def\vy{{\bm{y}}}
\def\vz{{\bm{z}}}

\def\mW{{\bm{W}}}

\DeclareMathAlphabet{\mathsfit}{\encodingdefault}{\sfdefault}{m}{sl}
\SetMathAlphabet{\mathsfit}{bold}{\encodingdefault}{\sfdefault}{bx}{n}

\def\sP{{\mathbb{P}}}
\def\sQ{{\mathbb{Q}}}

\newcommand{\normlp}{L^p}